\documentclass[journal=jacsat,manuscript=article]{achemso}

\usepackage{chemformula} % Formula subscripts using \ch{}
\usepackage[T1]{fontenc} % Use modern font encodings
\usepackage{subfigure}
\usepackage{float}
\usepackage{graphicx}% http://ctan.org/pkg/graphicx
\usepackage{array}% http://ctan.org/pkg/array
\usepackage{soul}
\author{Andrei Buin}
\affiliation[University of Guelph]
{Department of Mechanical Engineering, University of Guelph, 50 Stone Rd E, Guelph, ON, N1G 2W1}
\email{phquanta@gmail.com}
\author{Hung Yi Chiang}
\affiliation[University of Guelph]
{Department of Mechanical Engineering, University of Guelph, 50 Stone Rd E, Guelph, ON, N1G 2W1}

\author{S. Andrew Gadsden}
\affiliation[University of McMaster]
{Department of Mechanical Engineering, McMaster University, 1280 Main Street West
Hamilton, Ontario, Canada
L8S 4L7}
\email{gadsden@mcmaster.ca}

\author{Faraz A. Alderson}
\affiliation[University of Guelph]
{Department of Mechanical Engineering, University of Guelph, 50 Stone Rd E, Guelph, ON, N1G 2W1}

\title[An \textsf{achemso} demo]
  {De-novo Chemical Reaction Generation by Means of  Temporal Convolutional Neural Networks}

\abbreviations{IR,NMR,UV}
\keywords{American Chemical Society, \LaTeX}

\begin{document}

%\begin{tocentry}
%
%Some journals require a graphical entry for the Table of Contents.
%This should be laid out ``print ready'' so that the sizing of the
%Inside the \texttt{tocentry} environment, the font used is Helvetica
%8\,pt, as required by \emph{Journal of the American Chemical
%The surrounding frame is 9\,cm by 3.5\,cm, which is the maximum
%permitted for  \emph{Journal of the American Chemical Society}
%graphical table of content entries. The box will not resize if the
%content is too big: instead it will overflow the edge of the box.
%
%This box and the associated title will always be printed on a
%separate page at the end of the document.
%
%\end{tocentry}

\begin{abstract}
%    Recent advances of machine learning algorithms allow de-novo generation of data in variety of applications. While majority of these applications lie in video, audio and adjacent fields some
%limited applications to chemical drug generation using SMILES strings has been done. On the other hand, there are only  few works  are available  in the field of de-novo reaction generation.
We present here a combination of two networks, Recurrent Neural Networks (RNN) and Temporarily Convolutional Neural Networks (TCN) in de novo reaction generation using the novel Reaction Smiles-like representation of reactions (CGRSmiles) with atom mapping directly incorporated. Recurrent Neural Networks are known for their autoregressive properties and are frequently used in language modelling with direct application to SMILES generation. The relatively novel TCNs possess similar properties with wide receptive field while obeying the causality required for natural language processing (NLP). The combination of both latent representations expressed through TCN and RNN results in an overall better performance compared to RNN alone.  Additionally, it is shown that different fine-tuning protocols have a profound impact on generative scope of the model when applied on a dataset of interest via transfer learning.

\end{abstract}

\section{Introduction}
With advances in Deep Learning(DL)  generative methods, it is becoming more common to utilize DL's generative properties in a variety of applications. One such application is retorsynthetic planning, where given the products of a reaction, one tries to  predict reacting precursors that resulted in the products. There are works \cite{retro1,retrr2}  that are already using DL methods to guide retorsynthetic planning.
While it is a great tool for chemists, it is still lacking truly generative power when trying to generate novel reactions with unseen reaction centers and precursors.
Part of the problem is a lengthy language model describing chemical reactions in textual form, such as SMARTS/SMIRKS atom-mapped reaction representation. Only recently\cite{cgrnew} with the introduction of Condensed Graph of Reaction (CGR) \cite{cgrA} has complex reaction information  (reactants/products, bond formation/breaking) been successfully encoded into a simple textual representation.  In CGR, both reactants and products are combined into one single graph with bond creation and bond breaking incorporated, then expressed via SMILES-like strings.  We will refer to CGR as CGRSmiles throughout the paper.  One can tackle the task of generating SMILES-like strings via Recurrent Neural Networks(RNN)\cite{segler} with Long-Short-Term-Memory (LSTM) cells used to avoid problems of vanishing/exploding gradient.

On the other hand, a similar approach but with the use of Convolutional Neural Networks(CNN), namely  Temporal Convolutional Neural Networks (TCN)\cite{tcn1}  which use causal and dilated convolutions, have been used mostly in classification (prediction) applications.  These applications include text classification\cite{tcnApp7},  State-of-Charge battery estimation\cite{tcnApp1}, time series forecasting \cite{tcnApp2,tcnApp3,tcnApp5}, and protein-binding predictions\cite{tcnApp4}. By itself, TCNs have a generative power which comes from causal convolutions, and in this sense can be thought of as an alternative to RNN. Given that fact, there are virtually no  applications of novel DL architectures such as TCN in generative applied chemistry. By utilizing the TCN's generative power, we show that a combination of RNN-LSTM with TCN results in a better generative model compared to pure RNN models expressed as baseline models. In addition to this, SMILES generation previously and usually implied that the model learns SMILES independent of the context.  Context was only introduced via fine tuning a given dataset on which the model was subsequently refined.  We show that in-context SMILES generation exhibits more diverse structural motiff based on Tanimoto similarity scores, compared to the pure RNN-LSTM model without context.

Another contribution of this paper is the utilization  of a novel transfer learning protocol, which again is widely used in image/video applications\cite{google1} suitable for low-shot learning . With a traditional transfer learning protocol, where all the weights are "re-learned" on a particular dataset,  the model seems to  "memorize"  the particular dataset with application of learned grammar rules in initial training. As a result, it will try to generate reactions with a particular reaction template (reactants+products) as seen in a fine tuned dataset. We have observed this phenomena in other \cite{cgrnew} research, as well as our own. One can alleviate this problem by introducing an exhaustive reaction dataset for a particular problem, but this solution would not work in low-shot learning. With  a variant of the weight freezing, we show that our fine tuned  model, trained on our own dataset, significantly outperforms models that learned from an all-weights optimization transfer learning approach.

\section{Computational details}
The original work of Gupta et al. \cite{gupta} has been utilized and modified for RNN in parts of the code. As for TCN, we used a custom implementation of Remy \cite{KerasTCN}. All of the code was written in the Keras\cite{keras} framework with a Tensorflow\cite{tensorflow} backend. Custom vocabulary for CGRSmiles was incorporated with one-hot encodings.

RNNs with 2 to 3 Layers of stacked  LSTM cells were used as outlined in the results section. 512 hidden units were used for each RNN(LSTM) layer.
For the TCN network, 1 residual block was used without Normalization. We used a dilation vector of \[d=[1,2,4,8,16,32].\] A kernel size of 2 was used in 1D convolutional layers. 256 convolutional filters were used in the TCN residual block. For regularization, we have used a dropout of 0.5 for each of the LSTM and TCN layers. A softmax layer was used as the final layer for classification with categorical cross-entropy loss. No batch normalization was used in the TCN residual block.

For Seq2seq fingerprints (length of 768), an RNN of 3 stacked layers of (256)  GRU cells  with attention mechanism was used  \cite{seq2seq}. Additionally, Seq2Seq fingerprints were processed with Principal Component Analysis for dimensionality reduction. We kept 50 dimensions from PCA before it was fed  into a t-distributed stochastic neighbor embedding (tSNE)\cite{tsne}  analyzer. All SMILES manipulations, such as getting Tanimoto similarity scores and performing validity checks, were done using RDKit\cite{rdkit}.  For CGRSmiles generation, reaction center acquisition, and to/from reaction SMILES conversions along with aromaticity and valence checks, the CGRtools package \cite{cgr} was used. For the BiLSTM model \cite{bilstm}, the entire USPTO Dataset was cast as a classification problem by dividing the entire corpus into strings of 80 characters with a sliding window of 3, along with the 81st character being used as ground truth for the classifier.  We have tried longer character strings and shorter sliding windows, but this resulted in lengthy training times.

\subsection{Architectural design}
\subsection{Language Model}
When doing generative modeling in the context of Natural Language Processing (NLP), it is crucial to choose a proper language model. Usually RNN(LSTM) implies unidirectional context from past to future, whereas BiDirectional LSTM(BiLSTM) becomes inappropriate to use in NLP process. However, there are methods that use the BiLSTM\cite{bilstm} language model with certain adaptations. In one case, the entire dataset is represented as a single entire corpus, with the model trained to predict the next character given an N-length sequence in a sliding window manner. The problem with this approach is that  CGRSmiles have relatively long sequence lengths. Datasets created from the corpus of the original CGRSmiles reaction strings in the form of ${X,y}$, where $X={x_1,,…x_i,x_t}$ and label $y=x_{t+1}$,  becomes prohibitively large. As a result, training time also scales proportionally. Another Bidirectional adaptation involves interleaving BiDirectional sampling on the left and right of the sequence starting from the center character(BiMODAL)\cite{bimodal}. Additional augmentations of the adapted dataset, in this case, helps increase accuracy of the model. This again is at the expense of increased computational time. In general, the problem with Bidirectional language models, without proper adaptation, lies in the fact that the model has seen the whole context. If one, on the other hand, tries to use a vanilla BiLSTM with RNN-like training (where the target sequence is an original input sequence shifted 1 position to the right), the model will not learn to generate novel reactions, but instead just shift the input sequence to the right by 1 position.  These non-standard adaptations might require usage of dynamic graph neural computation rather than static graph computation due to graph modifications during runtime\cite{bimodal}. As a result, we adopt a standard language model suited for generating one token at a time given the left context, along with a refined fine tuning protocol.

\subsection{Model Training Protocol}
We used the Adam optimization algorithm\cite{adam} for training our models with  cross-entropy loss as the objective for optimization. For Seq2Seq and RNN, the original dataset split was $80\%$ for training and $20\%$ for test.  Learning rate was set to $10^{-3}$. Models were trained for 50 epochs using the original dataset and for 10 epochs in the ase of fine tuning. Batch size was 64 in the case of the original dataset, and 1 in the case of fine tuning.

\subsection{Fine Tuning Protocol}
A variety of fine tuning protocols were used. The original fine tuning protocol allowed all weights to be adjusted under the transfer learning approach. Another protocol involved freezing all layers except the last softmax layer. Additionally, we tried a decaying learning protocol for different layers \cite{fineTune1} using different learning rates for different layers, along with a different number of epochs trained for each layer.

\subsection{Sampling protocol}
Usually sampling involves the sampling softmax function: \[ \text{P}(y_i) = \frac{\exp(y_i)}{\sum_j \exp(y_j)}.\] This  gives the more syntactically correct, but less diverse, structures/reactions compared to the temperature controlled softmax function \[ \text{P}(y_i) = \frac{\exp(y_i/T)}{\sum_j \exp(y_j/T)},\] which has more diversity in generated structures/reactions but a smaller number of CGRSmiles strings. This sampling protocol has  been described elsewhere \cite{gupta,bimodal}.  GRSmiles strings were sampled at a sampling temperature of T = 0.7. For each analysis task, 30,000 CGRSmiles were generated.

\subsection{Datasets}
For the larger corpus we used the training dataset of Jin's  \cite{jin} USPTO atom-mapped dataset derived from Lowe's grant dataset \cite{lowe}. In the case of the fine tuning dataset, we used web scraped  reactions involving hydrogen peroxide($H_2O$ dataset) from PubChem \cite{pubchem} which later were preprocessed via Atom Mapper\cite{jaworski} for proper atom mapping. Jin's dataset and the $H_2O$ dataset were processed via the CGRtools library \cite{cgr} to obtain CGRSmiles strings. A max string of 156 characters was considered for  both datasets. After aromaticity and valence checks along with applying a max length of 156 characters for CGRSmiles strings, the larger dataset was reduced to 216,308 reactions. In the case of the smaller, fine tuning dataset, we acquired 166 atom-mapped reactions . It should be noted that most reactions in the smaller dataset are oxidation reactions (80\% oxidation reactions), meaning they contain O=O as part of the reactants.
%In generated dataset number of oxidation reactions is 92% meaning they contain O=O in reactant part

\subsection{Reaction Center and In-Context SMILES Analysis}
The most crucial part of any chemical reaction is the reaction center, i.e. atoms directly involved in bond creation/breaking.  To analyze novelty in terms of how the model performs, one needs a means of categorizing novelty, i.e. reaction centers. Fortunately, there is a way in the CGRtools library that encodes each substructural motif by hash function whose value is a  unique key. This value used  in categorizing known reaction centers within a dataset. Additionally, the hash value is used for the detection of novel reaction centers and comparison  between  known and unknown reaction centers. We do not categorize reactions based on known reaction centers, but instead with different 1st closest neighbor.   This is because the original dataset was not curated based on the presence of certain types of reaction centers, and all were considered. For in-context SMILES generation, each CGRSmile string was converted back to the reaction SMILES representation.  SMILES from the product and reactant parts were extracted for subsequent analysis using Tanimoto similarity scores.

\section{Results and discussion}

\begin{figure}[H]
  \begin{center}
      %\subfigure[]{\label{fig:Residiual} \includegraphics[width=0.48\textwidth]{Residiual.png}}

      %\subfigure[]{\label{fig:Baseline1}\includegraphics[width=0.35\textwidth]{Baseline1.png}}
      \subfigure[]{\label{fig:Baseline1} \includegraphics[width=0.26\linewidth,height=0.224\textheight]{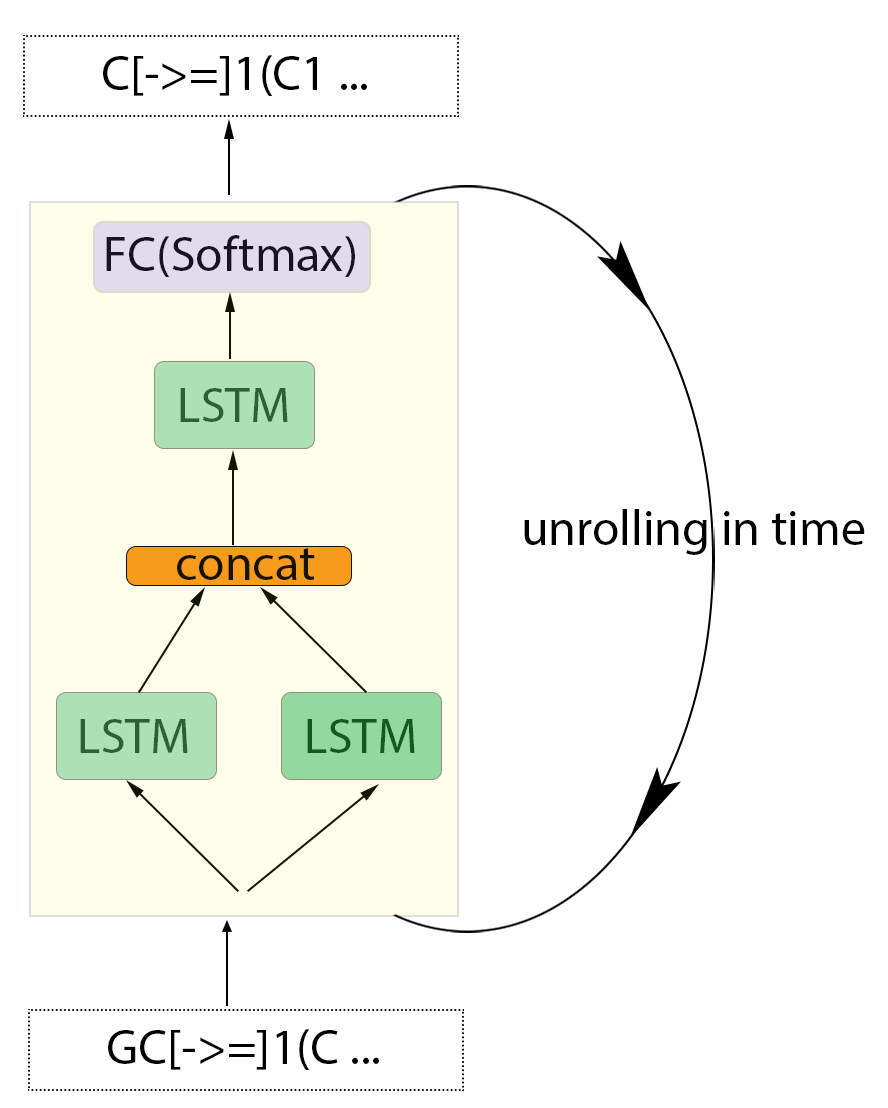}}
      %\subfigure[]{\label{fig:Baseline2}\includegraphics[width=0.35\textwidth]{Baseline2.png}}
      \subfigure[]{\label{fig:Baseline2} \includegraphics[width=0.26\linewidth,height=0.25\textheight]{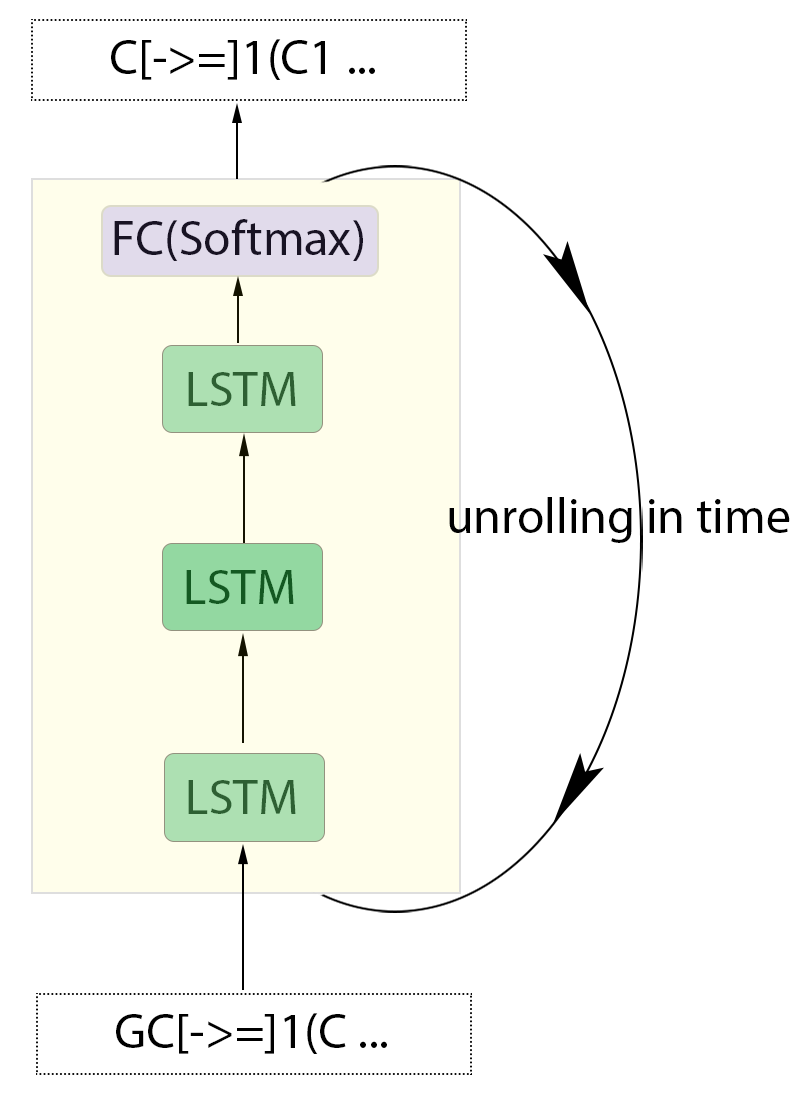}}
      \subfigure[]{\label{fig:Baseline3} \includegraphics[width=0.26\linewidth,height=0.25\textheight]{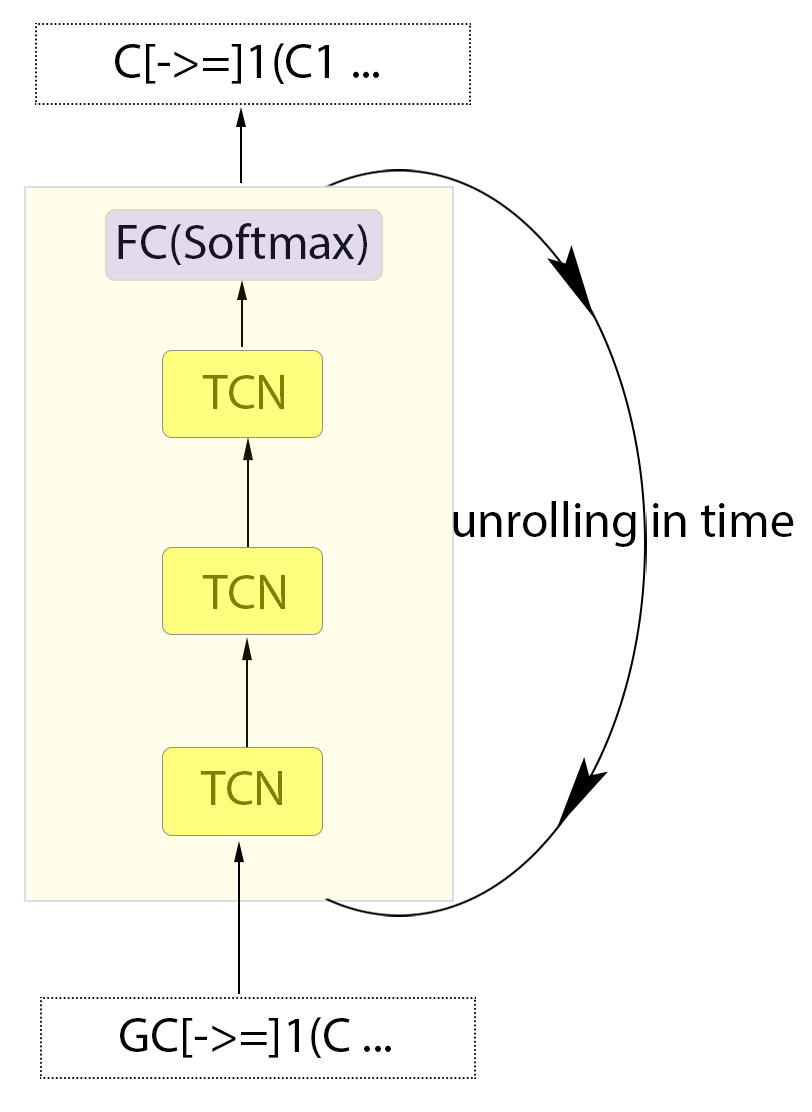}}
      \subfigure[]{\label{fig:Used} \includegraphics[width=0.26\linewidth,height=0.25\textheight]{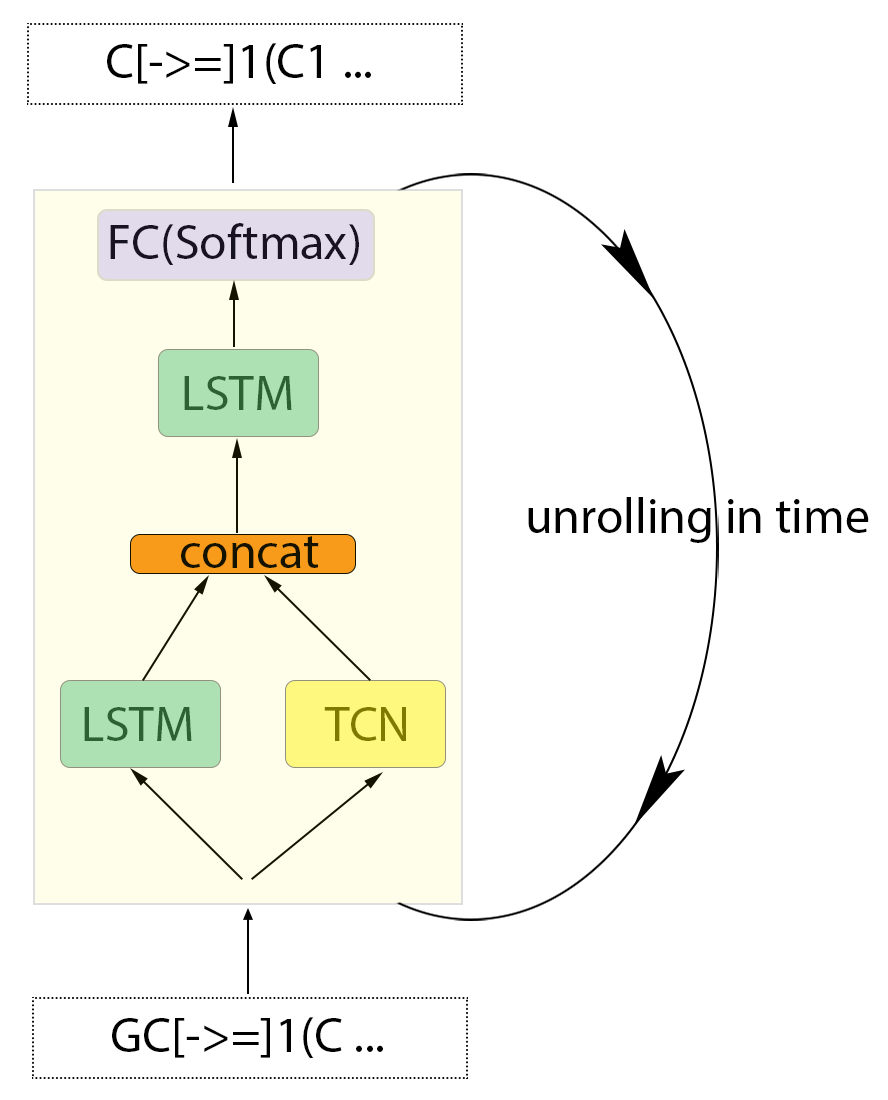}}
      \subfigure[]{\label{fig:Residiual} \includegraphics[width=0.4\linewidth,height=0.3\textheight]{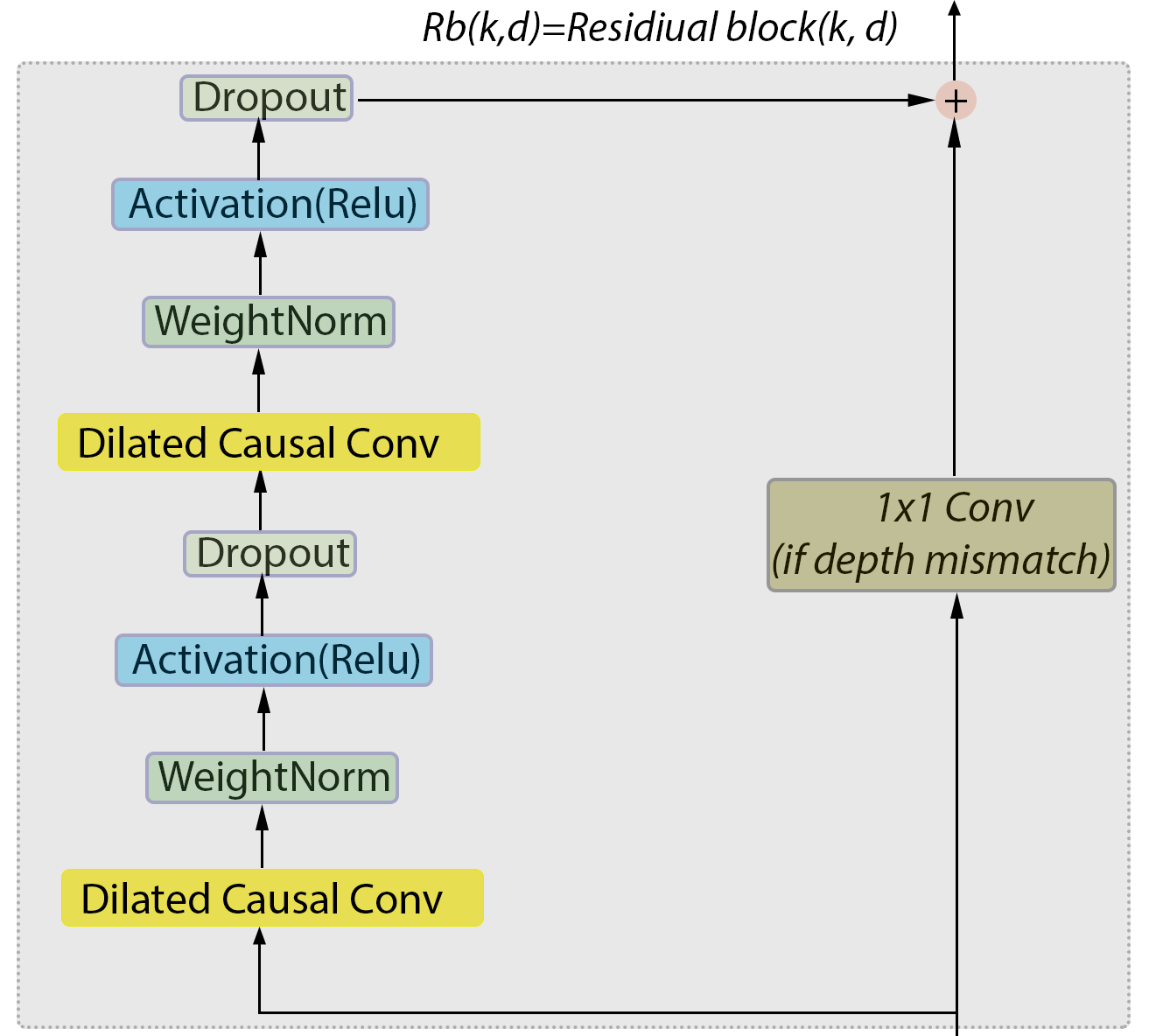}}
  \end{center}
  \caption{Baseline proposed architecture used (1) Baseline 1 (b) Baseline 2 (c) Baseline3 (d) Proposed (e) internals of TCN layer}
 \label{fig:Network}
\end{figure}

\begin{figure}[H]
  \begin{center}
      \subfigure[]{\label{fig:Baseline1} \includegraphics[width=0.45\linewidth,height=0.3\textheight]{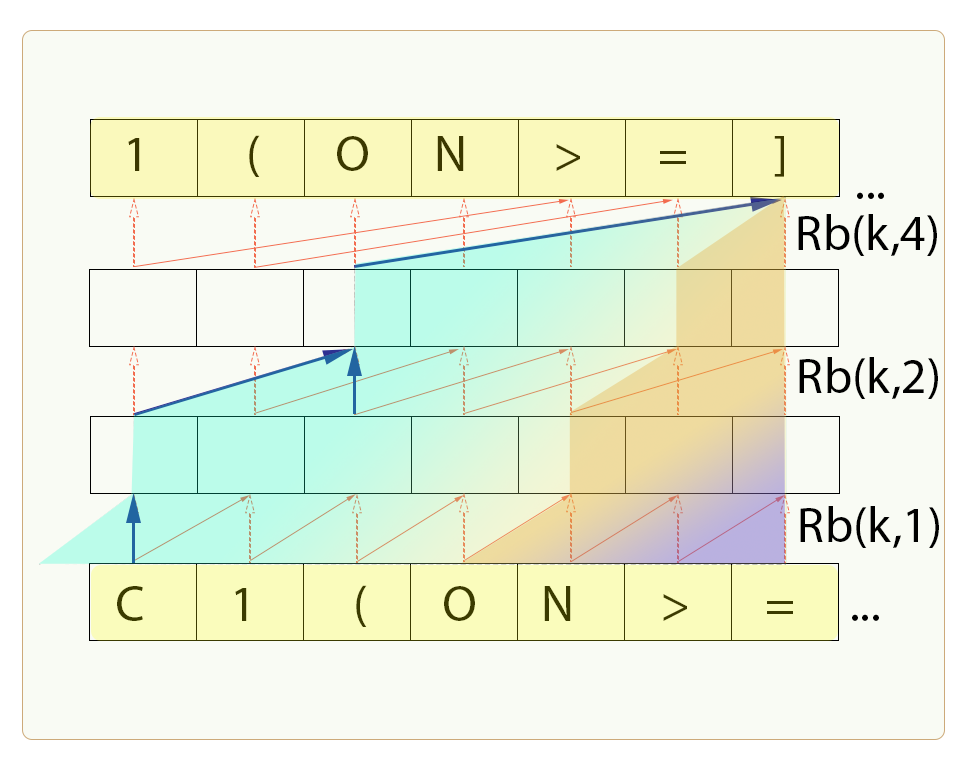}}
  \end{center}
  \caption{TCN with residual blocks and Receptive field of original convolution(light violet shading) and dilate convolution(light green shading)}
 \label{fig:RecField}
\end{figure}

Figure \ref{fig:Network} shows the Deep Learning architectures used. Baseline 1 - 3 are homogenous Deep Learning architectures with either TCN or LSTM layers used. The proposed architecture in Figure \ref{fig:Used} is, on the other hand, a combination of  both LSTM and TCN.  This architecture has the ability to learn from two latent representations. Figure \ref{fig:Residiual} shows a residual block, which is the core of TCN. Basically, TCN is a stack of residual blocks  which in turn consists of Dilated convolutional layers combined with Weight Normalization and Dropout layers.  This is shown in Figure \ref{fig:RecField}. In our case, no Weight Normalization was used and a dilation vector of \[d=[1,2,4,8,16,32]\] was utilized. 1x1 convolution is used in the case of depth mismatch between the input and output of the last dropout layer.
One could just utilize causal convolutions alone.  However, by introducing dilated convolutions, one has greatly enhanced the receptive field (i.e. past history) that TCN can look into. In other words, the last convolutional layer can see much further in the past when compared to plain causal convolutions, as seen in Figure \ref{fig:RecField}.  The receptive field, compared to vanilla convolutional operation (d=1) grows as \[ \frac{F^{TCN}} {F_{conv}} \propto  \frac{2^n} {n+1},\]
where n is the number of equivalent convolutional and residual layers of TCN with exponentially growing dilation ($2,4,8....$). Clearly, TCN has the advantage. As demonstrated in Figure \ref{fig:RecField}, this advantage is 2. In our case, the actual advantage is roughly 9 - fold.

Next, we considered how well the models perform when generating novel reactions based on initial training. From table \ref{tab:table1}, one can see that all of the models give a relatively high number of unique and valid CGRSmiles strings.  However, both TCN and TCN+RNN(LSTM) both gave significantly higher numbers of novel reaction centers.  For comparison, we also compared our results with the BiLSTM language model \cite{bilstm} that has 2 BiLSTM layers with 128 hidden units each. Results are shown in table \ref{tab:table1}. One can see that the number of valid CGRSmiles string is significantly lower for the BiLSTM model, while the amount of reaction centers is higher compared  to the TCN and TCN+RNN models. One aspect to note is that the training time required for BiLSTM is significantly higher compared to both the TCN and TCN+RNN models. A good compromise is achieved via the combination of TCN+RNN, where the number of valid CGRSmiles strings was the highest among all models and the number of novel reaction centers was relatively high, albeit not the highest.

\begin{table}
\begin{tabular}{ |p{5cm}||p{2.5cm}||p{2.5cm}|p{2.5cm}|}
 \hline
  Model & Valid(\%) & Unique(\%)  & N(RC)    \\
 \hline
 Baseline1   & 93.42     & 98.84 & 877  \\
 Baseline2   & 94.40 & 99.3     & 873 \\
 TCN    & 85.52 & 98.2    & 1943 \\
 TCN+RNN(LSTM)      & 94.71 & 98.66    & 1239  \\
 BiLSTM(80,3) & 78.52 & 99.87    & 2606  \\
 Dataset              & N/A  & N/A     & 12308 \\
\hline
\end{tabular}
\caption{\label{tab:table1} Generative properties of a variety of models.}
\end{table}

\begin{figure}[H]
  \begin{center}
      \subfigure[]{\label{fig:aaa} \includegraphics[width=0.66\textwidth]{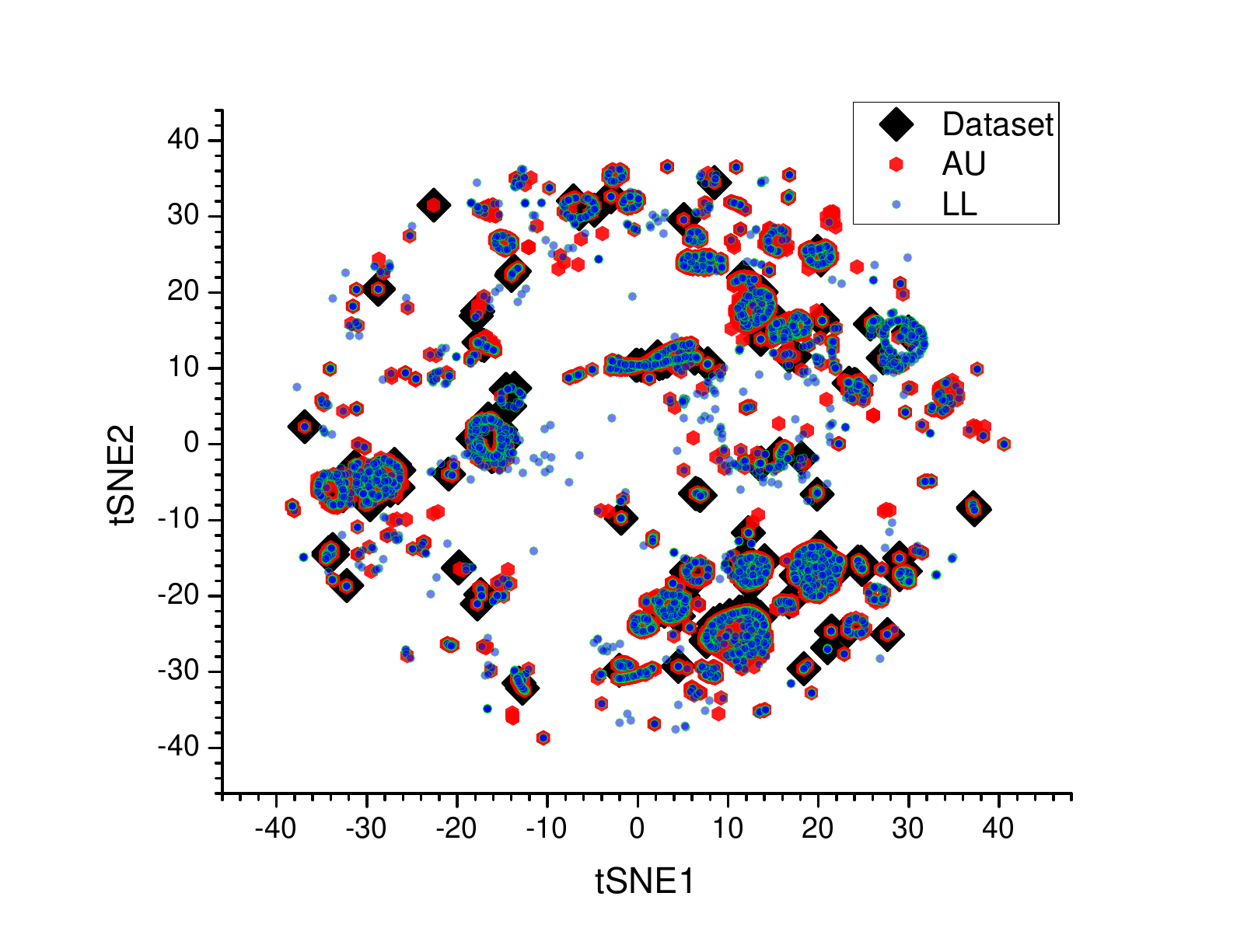}}
  \end{center}
  \caption{tSNE plots of generated CGRSmiles compared to fine-tuned dataset and LL protocol where all layers are unfrozen.}
 \label{fig:TSNE}
\end{figure}

The next step was to explore a variety of fine tuning protocols along with in-context SMILES generation. Table \ref{tab:table2} shows that if one is to allow the model to freely adapt to a smaller dataset with all weights being adjustable (all unfrozen, AU in Table \ref{tab:table2}), "memorization" or overfitting of the model on the novel dataset will occur. On the other hand and while keeping only the last layer unfrozen (LL in Table \ref{tab:table2}), the model is capable of transferring its knowledge from previous learning more efficiently. Other fine tuning protocols have been tried as well, one of which is shown in table \ref{tab:table2} (P1, or Protocol1) and provides results lying between the 2 extremes from the other two protocols.  Interestingly enough, the model with the LL-transfer learning approach was able to sample CGRSmiles with Na, Pt, and Se ions in them.  These ions were not a part of the smaller dataset, but the model has seen some examples of such reactions in the larger dataset with Na, Pt, Se ions and applied its knowledge during the fine-tuning phase.
\begin{table}
\begin{tabular}{ |p{5cm}||p{2.5cm}||p{2.5cm}|p{2.5cm}|}
 \hline
  Model & Valid(\%) & Unique(\%)  & N(RC)    \\
 \hline
 AU  & 98.65     & 9.98 & 63  \\
 P1  & 94.95     & 21.12 & 97  \\
 LL  & 91.92     & 60.40 & 288  \\
 $H_2O_2$ Dataset     & N/A  & N/A     & 64 \\
\hline
\end{tabular}
\caption{\label{tab:table2} Fine Tuning Properties of a variety of Fine-Tuning Protocols for TCN+RNN. AU - All Unfrozen, P1 - Protocol1 where layers= [[1,2], 4, 5] were trained for different numbers of epochs [2,5,10] with learning rates=[$10^{-6},10^{-5},5*10^{-4}$] in sequential order starting from last layer. LL - only Last Layer (SoftMax) was unfrozen, while all other layers were frozen.}
\end{table}
We also computed tSNE plots of generated CGRSMiles as shown in Figure \ref{fig:TSNE}. One can see that the results are in agreement with Table \ref{tab:table2} as expected, with Last Layer(LL), unfrozen upon transfer-learning,  giving the highest amount of different reaction centers.   This indicates the possibility of few-shot learning with only a few samples from the smaller dataset.

\begin{figure}[H]
  \begin{center}
      \subfigure[]{\label{fig:PictureReactionSmiles} \includegraphics[width=1.00\linewidth,height=0.2\textheight]{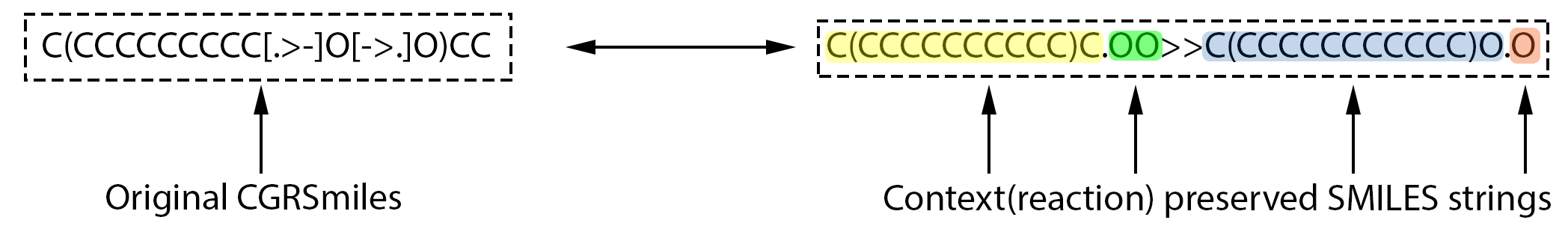}}
       \subfigure[]{\label{fig:Tanimoto2} \includegraphics[width=0.46\linewidth,height=0.254\textheight]{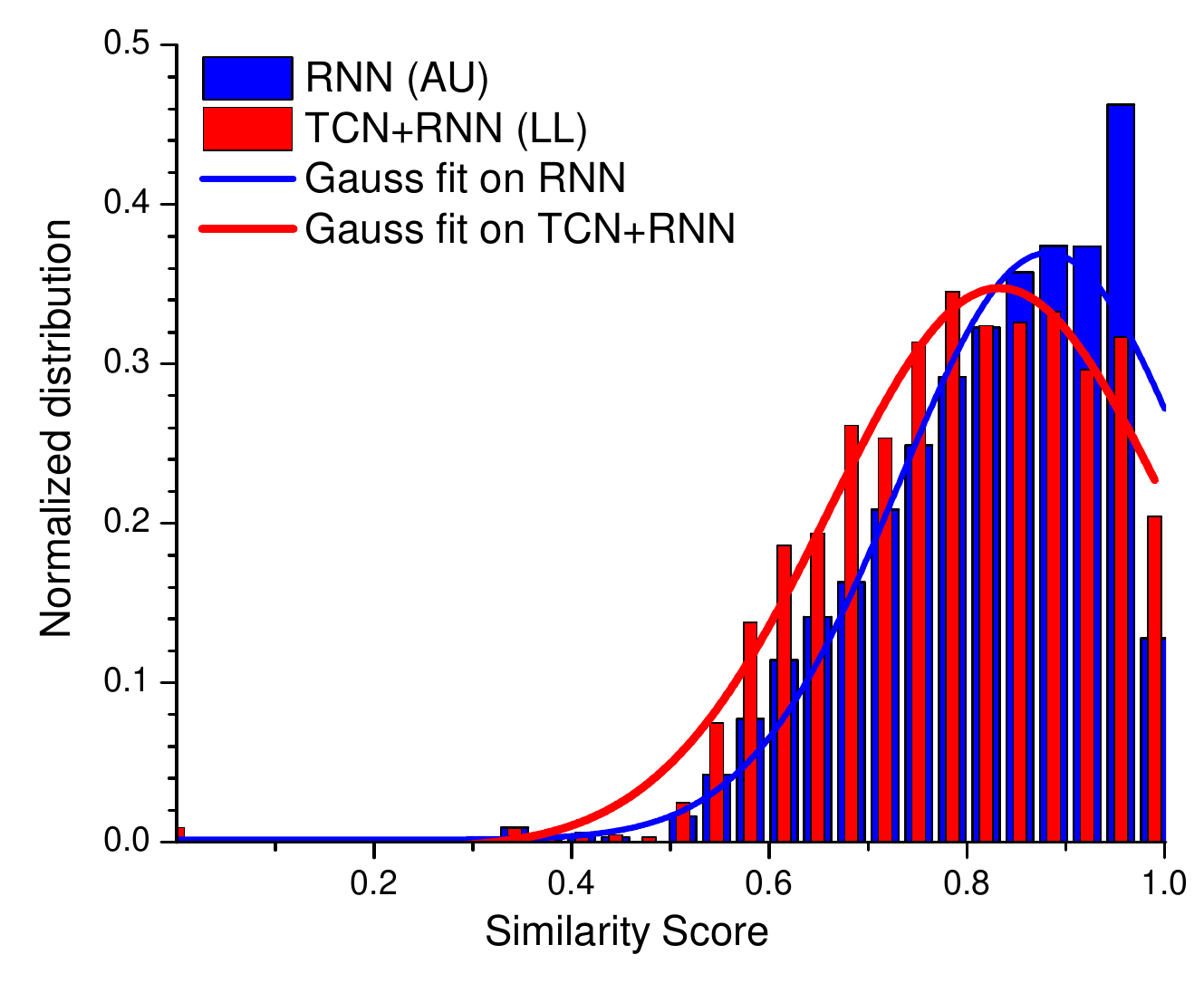}}
      \subfigure[]{\label{fig:Tanimoto1} \includegraphics[width=0.46\linewidth,height=0.254\textheight]{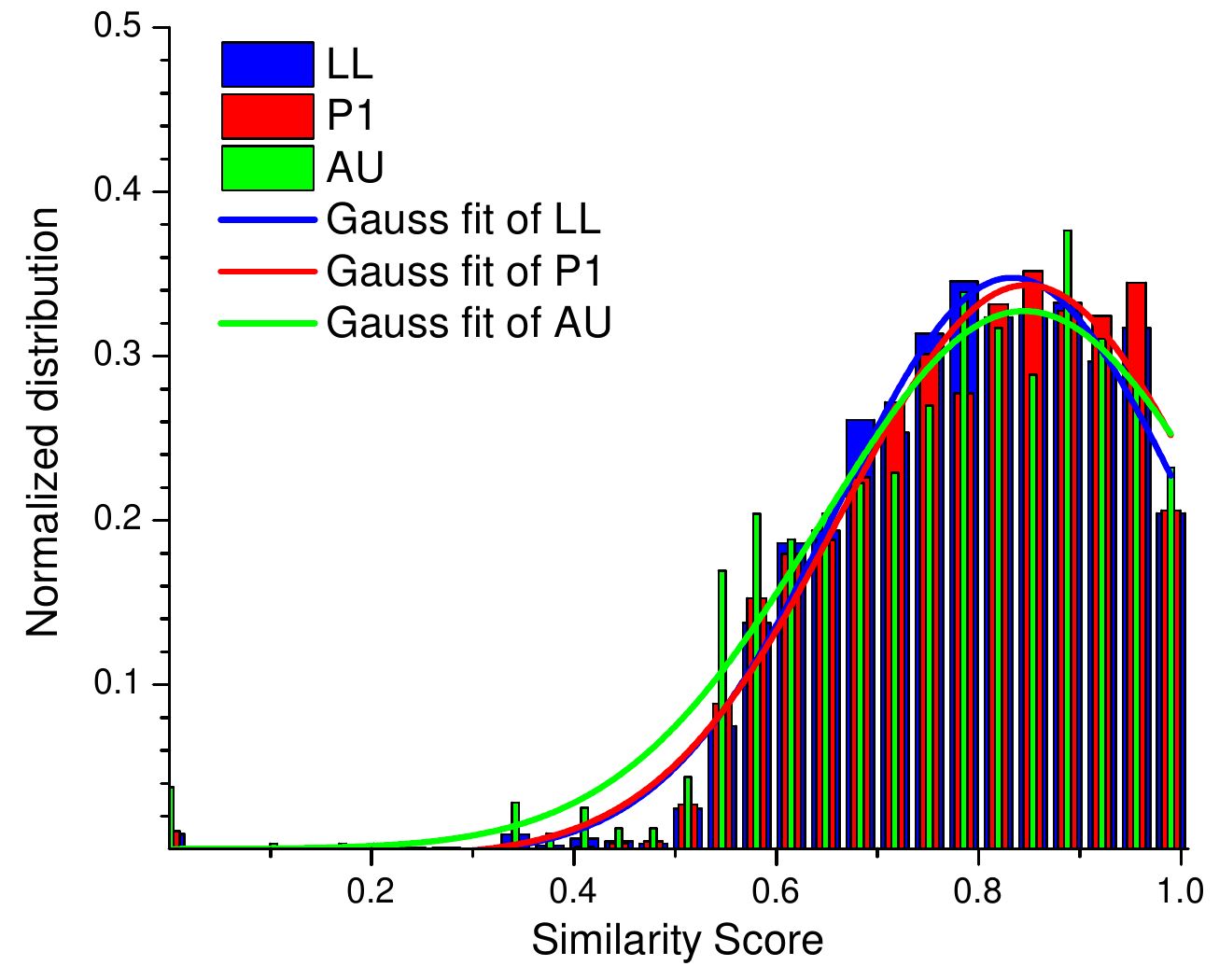}}

  \end{center}
  \caption{(a) Mapping between CGRSmiles and SMIRKS representation. Please note that during conversion, atom mapping is lost. (b) InContext Tanimoto Scores for RNN and TCN+RNN. (c) Dependence of Tanimoto score in case of TCN+RNN with respect to fine-tuning protocols.}
 \label{fig:InContext}
\end{figure}

To explore the SMILES of reactants and products that participated in a given reaction , we performed the conversion of CGRSmiles back to reaction SMIRKS consisting of the participating compounds. In this case however, the information about atom mapping is lost during the conversion and only reactants and products are preserved. Figure \ref{fig:PictureReactionSmiles} shows a typical example of such conversion. In this case, each reactant/product is highlighted with different colors. By analyzing these participating chemical formulations, one can look into the context that created those SMILES strings. In other words, SMILES extracted from SMIRKS representation are not mere SMILES, but rather SMILES with context in them, since each SMILES string in this case is tied with reaction. In a broader sense, it means that some reactants and products are more common than others in certain reactions.  Figure \ref{fig:Tanimoto2} and Figure \ref{fig:Tanimoto1} show a difference in in-context SMILES similarity scores in the case of RNN and TCN+RNN. One can see that TCN+RNN gives a more diverse SMILES scaffold compared to RNN by having a smaller mean Tanimoto score. Please note that both were compared at different fine-tuning protocols: one is done at AU, and another is done at LL. However, Figure \ref{fig:Tanimoto1} shows that fine tuning protocol has a little effect on in-context-SMILES generated strings and as a result the shift towards lower Tanimoto scores could be attributed to architectural choice.

\begin{figure}[H]
  \begin{center}
      \subfigure[]{\label{fig:R1} \includegraphics[width=0.86\textwidth]{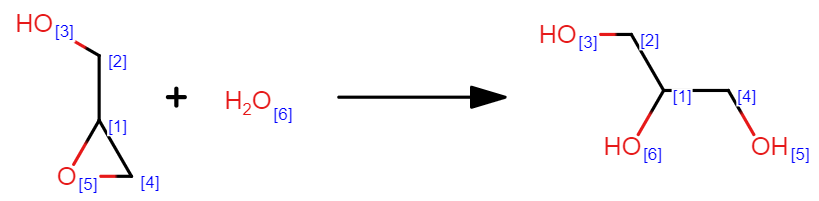}}
      \subfigure[]{\label{fig:R2} \includegraphics[width=0.86\textwidth]{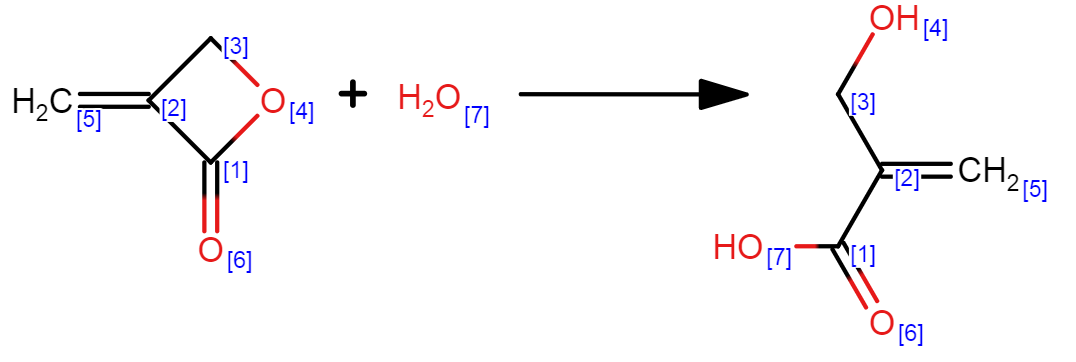}}
      \subfigure[]{\label{fig:R3} \includegraphics[width=0.86\textwidth]{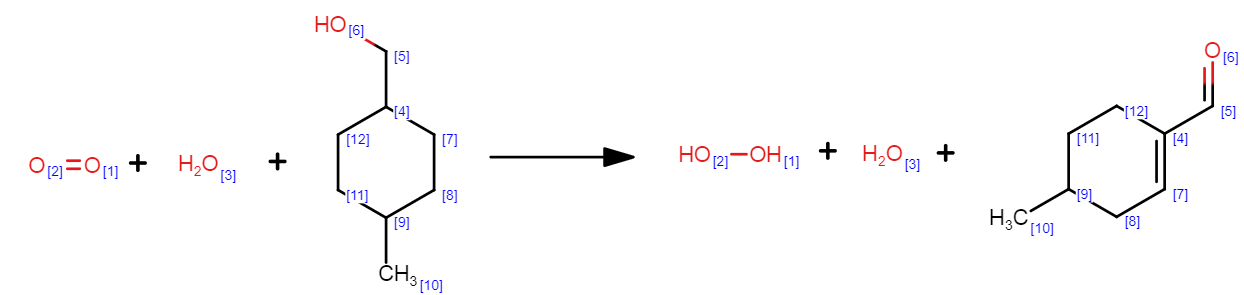}}
      \subfigure[]{\label{fig:R4} \includegraphics[width=0.66\textwidth]{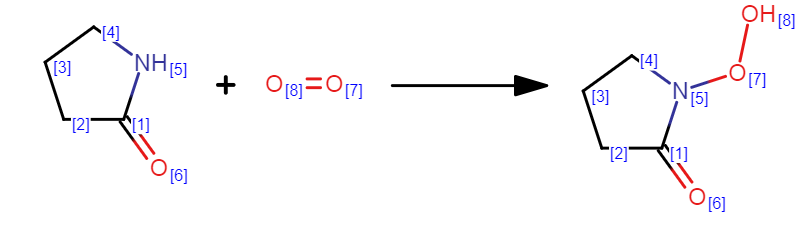}}
  \end{center}
  \caption{A variety of Generated atom-mapped reactions with novel and known reaction centers. }
 \label{fig:Reactions}
\end{figure}

Finally, to explore generative capabilities, we studied some of the reactions generated by the models. Out of all the Fine-Tuning protocols, LL gave the most reactions with novel reaction centers. In total, LL gave on the order of 800 reactions with novel RC, whereas all the other protocols gave only reactions on the order of 100 with novel RC. Figure \ref{fig:R1} shows an example of an unseen reaction, not present in the dataset, with novel RC. This is glycidol hydrolysis, with no mistakes\cite{R1,R1_1}.  Another reaction with novel RC is shown in Figure \ref{fig:R2}. The closest reaction  to this one is diketene hydrolysis \cite{R2,R2_1}. Interestingly enough, the model is able to learn how to open the ring, albeit with some errors such as the wrong placement of the $CH_2$ group and O. Another reaction, but with known reaction center this time, is the oxidation of the cyclohexanol derivative shown in Figure \ref{fig:R3}. This is a feasible pathway of oxidation for the cyclohexanol derivative, as the closest reaction is the oxidation of cyclohexanol\cite{R3} with a similar pathway. In our case, oxidation was done in presence of water, whereas the cited work \cite{R3} uses tert-Butyl hydroperoxide (TBHP) as an oxidizing agent. This phenomenon can be attributed  to the fact that most reactions in the fine-tuning  dataset are oxidation reactions (80$\%$ oxidation reactions as mentioned previously), meaning that they contain O=O in the reactants. For generated CGRSmiles, the number of oxidation reactions is 92$\%$, whereas the rest of the reactions contain mainly $H_2O$ and $H_2O_2$ as precursors. In addition to this, the collected dataset involving $H_2O_2$ did not contain metadata attributed to reactions such as  catalyzers, oxidizing agents etc. Instead, only plain SMIRKS reaction representations were collected from open sources. In addition to this, reaction \ref{fig:R3} has an invalid stoichiometry. Most errors of this type involve a wrong number of implicit hydrogens. We have  observed phenomena of reactions being unbalanced in line with other work\cite{cgrnew}, mostly in the imbalance of implicit hydrogens. This has been attributed\cite{cgrnew} to the USPTO reaction database being imbalanced in the first place. Additionally, a small portion (2.5$\%$ of all generated smiles) of  errors, such as copying the reactant part directly into the product part, was also observed. The reaction shown in Figure \ref{fig:R4} is a reaction with novel RC -  an initial pathway for gold reduction by 2-pyrrolidinone \cite{R4} with the wrong placement of the $O-OH$ group. However, one should keep in mind that the original work\cite{R4_1} cited by Li\cite{R4} et al. uses Nuclear Magnetic Resonance (NMR) shifts of $^{13}C$ in determining the structure of the intermediate compound and the $N$ atom has a methyl group. This is in contrast to the work of Li\cite{R4} et. al and our work, where there is no methyl group attached to $N$, and resulting shifts do not necessarily correspond to the $NH$ group. In other words, the placement of the $-OOH$ group does not necessarily have to be on a carbon atom as no rigorous NMR analysis was done in the case of Li\cite{R4} et al.
\section{Conclusion}
This work presents a step forward towards unsupervised de-novo reaction generation. The contribution of this work is threefold: First, it explores an alternative TCN Deep Learning architecture in comparison with RNN by itself. Second, it is shown that this approach allows for context-aware SMILES generation. Lastly, it is shown that fine-tuning protocols have a significant contribution towards chemical domain adaptation in the chemical space, which in turn enables few-shot learning on smaller datasets upon transfer learning. The model with the best fine-tuning protocol was able to discover reactions which it has not seen, but was present in previous published work. This leads to the possibility of gaining reaction insight before the synthesis stage.

%IUUI

%reaction to generate:
%1. C(C([O-])=O)(CSC=CCC(=O)N)[NH3+].O.O=O>>OO.S(C=CCC(=O)N)(CC(C([O-])=O)[NH3+])=O

%Mention NADE approach as bidirectional to Sequence generation

%IT is feasible to utilize Bidirectionals lstm for structure generation as work of see reference (cite) -

\section{Data and Software Availability}
Availability of data and materials CGRSmiles Dataset and Collected Hydrogen Peroxide datasets, along with Generated CGRSmiles and Python scripts are available at:     
\url{https://github.com/phquanta/CGRSmiles.git}

\begin{acknowledgement}

%Please use ``The authors thank \ldots'' rather than ``The
%authors would like to thank \ldots''.
%
%The author thanks Mats Dahlgren for version one of \textsf{achemso},
%and Donald Arseneau for the code taken from \textsf{cite} to move
%citations after punctuation. Many users have provided feedback on the
%class, which is reflected in all of the different demonstrations
%shown in this document.

\end{acknowledgement}


\bibliography{bib1}

\end{document}